\documentclass[10pt,twocolumn,letterpaper]{article}
\usepackage[pagenumbers]{iccv} 

\definecolor{iccvblue}{rgb}{0.21,0.49,0.74}
\usepackage[pagebackref,breaklinks,colorlinks,allcolors=iccvblue]{hyperref}

\usepackage{ulem}
\usepackage{booktabs}
\usepackage{amssymb}

\usepackage{pifont}       %
\usepackage{bbding}       %
\usepackage{fontawesome}
\usepackage{xspace}
\usepackage{microtype}
\usepackage{graphicx}
\usepackage{booktabs} %
\usepackage{subcaption}
\usepackage{amsmath}
\usepackage{amssymb}
\usepackage{mathtools}
\usepackage{amsthm}
\usepackage{multirow}
\usepackage{multicol}
\usepackage{makecell}
\usepackage{colortbl}
\usepackage{xcolor}
\usepackage{float}
\usepackage{fixltx2e}
\usepackage{wrapfig}

\definecolor{darkblue}{rgb}{0.121, 0.316, 0.590}

\definecolor{green}{HTML}{009000} 
\renewcommand{\paragraph}[1]{\vspace{2pt}\noindent\textbf{#1}}
\newlength\savewidth\newcommand\shline{\noalign{\global\savewidth\arrayrulewidth
  \global\arrayrulewidth 1pt}\hline\noalign{\global\arrayrulewidth\savewidth}}
\newcolumntype{x}[1]{>{\centering\arraybackslash}p{#1pt}}
\newcolumntype{y}[1]{>{\raggedright\arraybackslash}p{#1pt}}
\newcolumntype{z}[1]{>{\raggedleft\arraybackslash}p{#1pt}}
\definecolor{mygreen}{RGB}{0, 205, 108}
\definecolor{convcolor}{HTML}{412F8A}
\definecolor{resnetcolor}{HTML}{8DA0CB}
\definecolor{vitcolor}{HTML}{fc8e62}

\definecolor{mygreen}{RGB}{0, 205, 108}
\definecolor{code0}{RGB}{223, 227, 241}
\definecolor{code1}{RGB}{250, 219, 223}
\definecolor{convcolor}{HTML}{412F8A}
\definecolor{resnetcolor}{HTML}{8DA0CB}
\definecolor{vitcolor}{HTML}{fc8e62}

\usepackage{pifont}       %
\usepackage{bbding}       %
\usepackage{fontawesome}
\usepackage{xspace}

\hypersetup{
    colorlinks=true,
    linkcolor=red,
    citecolor=blue,
    filecolor=magenta,      
    urlcolor=magenta,
    }
\newcommand{\tablestyle}[2]{\setlength{\tabcolsep}{#1}\renewcommand{\arraystretch}{#2}\centering\footnotesize}

\definecolor{orange}{HTML}{ff7f0e}
\definecolor{blue}{HTML}{1f77b4}
\definecolor{baselinecolor}{gray}{.9}
\newcommand{\baseline}[1]{\cellcolor{baselinecolor}{#1}}

\def\paperID{15373} 
\def\confName{ICCV}
\def\confYear{2025}

\newcommand{\cmark}{{\color{teal}\ding{51}}}
\definecolor{crimsonred}{RGB}{220,20,60}
\definecolor{maroon}{cmyk}{0,0.87,0.68,0.32}
\newcommand{\xmark}{{\color{crimsonred}\ding{55}}}

\title{ORAL: Prompting Your Large-Scale LoRAs via Conditional Recurrent Diffusion}

\author{Rana Muhammad Shahroz Khan\textsuperscript{1} \quad 
  Dongwen Tang\textsuperscript{2} \quad
  Pingzhi Li\textsuperscript{1} \quad
  Kai Wang\textsuperscript{2} \quad
  Tianlong Chen\textsuperscript{1}\\
  {\small \textsuperscript{1}The University of North Carolina at Chapel Hill\quad
  \textsuperscript{2}National University of Singapore} \quad \\ 
}

\begin{document}
\maketitle
\begin{abstract}
Parameter generation has emerged as a novel paradigm for neural network development, offering an alternative to traditional neural network training by synthesizing high-quality model weights directly. In the context of Low-Rank Adaptation~(LoRA) for evolving (\textit{i.e.}, constantly updated) large language models~(LLMs), this approach promises efficient adaptation without costly retraining. However, existing methods face critical limitations in simultaneously achieving scalability and controllability. In this paper, we introduce \texttt{ORAL}, a novel \textbf{conditional recurrent diffusion} framework that addresses these challenges. \texttt{ORAL} incorporates a novel conditioning mechanism that integrates model architecture and textual task specifications, enabling the generation of task-specific LoRA parameters that can seamlessly transfer across evolving foundation models. Our approach successfully scales to billions-of-parameter LLMs and maintains controllability. Through extensive experiments across seven language tasks, four vision tasks, and three multimodal tasks using five pre-trained LLMs, we demonstrate that \texttt{ORAL} generates high-quality LoRA parameters that achieve comparable or superior performance to vanilla trained counterparts. 

\end{abstract}    
\section{Introduction}

Recent advancements in generative AI have been fueled by the vast amount of valuable data throughout the Internet, demonstrating profound transformations in the creation of texts, images, and videos~\citep{liu2024deepseek,achiam2023gpt,zhao2024model,zhang2024mm15methodsanalysis,lin2023video,liu2023visualinstructiontuning,videoworldsimulators2024}. Inspired by these breakthroughs, researchers have begun to investigate whether the vast number of online pre-trained model checkpoints can serve as an untapped resource for AI development. Building on weight-space learning~\citep{eilertsen2020classifying,schurholt2021self,zhao2023improving,horwitz2024learningmodelweightsusing}, which treats neural network parameters as a distinct modality, parameter generation has emerged as a novel paradigm.

Pioneering works in this direction, such as P-Diff~\citep{wang2024neuralnetworkdiffusion}, have demonstrated the potential of diffusion models to generate neural network parameters that match or even surpass the performance of conventionally trained networks. P-Diff primarily focuses on unconditional generation, treating parameter synthesis as a distribution modeling problem without explicit control mechanisms. While this approach has shown promising results for small-scale architectures, it lacks the ability to guide the generation process toward specific tasks or model architectures, limiting its practical applications in rapidly evolving LLM ecosystems.

\begin{table}[t]
\centering
\begin{tabular}{lccc}
\toprule
\midrule
Method & Scalable & Controllable & Portable \\
\midrule
P-Diff~\citep{wang2024neuralnetworkdiffusion} & \xmark & \xmark & \xmark \\
Cond P-Diff~\citep{jin2024conditionalloraparametergeneration} & \xmark & \cmark & \xmark \\
RPG~\citep{wang2025recurrentdiffusionlargescaleparameter} & \cmark & \xmark & \xmark \\
\texttt{ORAL} (Ours) & \cmark & \cmark & \cmark \\
\midrule
\bottomrule
\end{tabular}
\caption{Comparison of LLM parameter generation methods across three dimensions: \textit{Scalability} (ability to generate parameters at the scale of hundreds of millions); \textit{Controllability} (support for task-specific conditional generation); and \textit{Portability} (capacity to adapt to evolving foundation models without retraining). }
\label{tab:comparison}
\end{table}

A subsequent work, Cond P-Diff~\citep{jin2024conditionalloraparametergeneration}, introduced conditionality to parameter generation, enabling task-specific adaptation by incorporating textual guidance. This significant advancement allows the synthesis of Low-Rank Adaptation~(LoRA) parameters tailored for particular downstream tasks. However, Cond P-Diff faces critical limitations in: ($1$)~scaling, typically constrained to generating around one million parameters, and ($2$)~flexibility to adapt to weight changes in foundation models, requiring costly retraining whenever the model evolves.

The recent RPG~\citep{wang2025recurrentdiffusionlargescaleparameter} framework alleviates the scalability challenge through a recurrent diffusion architecture. Despite this breakthrough in scale, RPG focuses exclusively on unconditional generation, offering no mechanisms to guide the generation process toward specific tasks or adaptation requirements. This is a significant limitation for practical deployment scenarios where task specificity is crucial.

Meanwhile, the landscape of LLMs undergoes frequent updates to their parameters~\citep{khan2024portllm}, necessitating parameter generation methods that can efficiently adapt to these changes without the need for complete retraining. As summarized in Table~\ref{tab:comparison}, existing methods have only partially addressed the challenges in LLM parameter generation. While Cond P-Diff enables controllability and RPG achieves scalability, no current approach successfully combines all three crucial properties: \textit{scalability}, \textit{controllability}, and \textit{portability} across evolving foundation model. These challenges lead us to our center research question:


\textit{\uline{RQ:} How can we design a parameter generation framework that flexibly adapts to continuously evolving foundation models while maintaining efficient scaling properties for practical deployment?}

\begin{figure*}[htbp]
    \vspace*{0pt}
    \centering
    \includegraphics[width=0.93\textwidth]{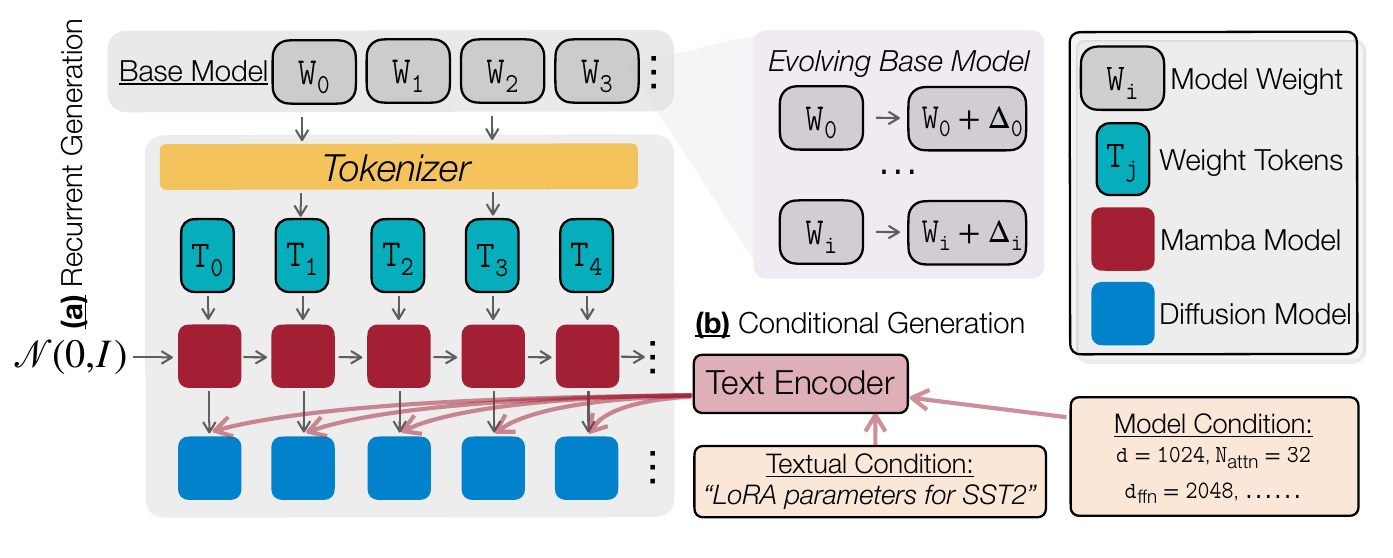}
    \vspace{-10pt}
    \caption{\textbf{Overview of our ORAL framework}. \uline{\textbf{(a)} Recurrent Generation}: The foundation model weights are processed through a tokenizer to create weight tokens, which are then fed into a recurrent architecture consisting of both Mamba and Diffusion models to generate parameters from noise $\mathcal{N}(0,I)$. \uline{\textbf{(b)} Conditional Generation}: Our approach supports \textbf{evolving} foundation model by adapting parameters $\mathtt{W}_i$ to updated foundation models $\mathtt{W}_i + \Delta_i$ without retraining, which is enabled by our novel conditioning mechanism that incorporates model architecture and textual task specifications.}
    \vspace{-10pt}
    \label{fig:routing-frequency}
\end{figure*}

We introduce \texttt{ORAL}, a novel framework that enables evolvable and scalable LoRA parameter generation. Our approach, building upon the recurrent diffusion backbone introduced in RPG~\citep{wang2025recurrentdiffusionlargescaleparameter}, proposes a novel conditioning mechanism that incorporates both model architectural and textual specifications of adaptation characteristics as inputs to the diffusion process. Specifically, Our contributions are:

\begin{itemize}
\item We introduce a novel conditioning mechanism that integrates both model architectural and textual task specifications, enabling flexible generation of LoRA parameters for specific downstream tasks.
\item We develop a novel conditional parameter generation pipeline that facilitates seamless transfer of generated LoRA parameters to evolving foundation models without requiring resource-intensive retraining, ensuring compatibility with rapidly advancing foundation models.
\item Through comprehensive experiments across seven language tasks, four vision tasks, three multimodal tasks, and five pre-trained LLMs, we demonstrate the efficacy of our approach. The results validate that \texttt{ORAL} achieves remarkable scaling efficiency, effectively handling models with up to $7$B parameters while maintaining or exceeding the performance of traditional fine-tuning approaches.
\end{itemize}

\section{Related Works}
\paragraph{Diffusion Models.} Diffusion models have transformed generative AI, successfully creating high-quality images and other complex data. These methods~\citep{NEURIPS2021_49ad23d1,hertz2022prompt,li2022efficient, li2023your} build on non-equilibrium thermodynamics principles~\citep{jarzynski1997equilibrium, pmlr-v37-sohl-dickstein15, peebles2023scalable} and have followed a development path similar to GANs~\citep{brock2018large, isola2017image, zhu2017unpaired, goodfellow2014generative}, VAEs~\citep{kingma2013auto, NEURIPS2019_5f8e2fa1}, and flow-based models~\citep{dinh2014nice, pmlr-v37-rezende15}. They've evolved from simple unconditional generation to sophisticated systems that respond to text, images, and other structured inputs. Research on diffusion models spans four main areas: The first focuses on improving visual quality—exemplified by DALL-E 2~\citep{ramesh2022hierarchical}, Imagen~\citep{saharia2022photorealistic}, and Stable Diffusion~\citep{rombach2022high, podell2023sdxl}—significantly enhancing the realism of generated outputs. The second addresses their initially slow sampling process through methods like DDIM~\citep{song2020denoising}, Analytic-DPM~\citep{bao2022analytic}, and DPM-Solver~\citep{NEURIPS2022_260a14ac}, which have dramatically accelerated generation speeds. The third area involves theoretical work reexamining diffusion through continuous mathematical frameworks~\citep{feng2023score, NEURIPS2019_3001ef25, salimans2022progressive, shi2025make}, particularly score-based generative modeling~\citep{feng2023score}, deepening our understanding of these techniques. The fourth focuses on applying diffusion across varied domains like Audio~\citep{kong2020diffwave}, 3D Point Generation~\citep{Luo_2021_CVPR} and Anomaly Detection~\citep{wolleb2022diffusion}, demonstrating their versatility as a generative approach.

\paragraph{Conditional Generation.} The area of conditional generation has experienced incredible progress over the last decade, evolving through several distinct methodological stages. Initial efforts focused on conditional GANs~\citep{mirza2014conditional,8100115,zhu2017unpaired, Ho2020DenoisingDP}, which pioneered the integration of specific control signals into the generative process. Although these methods showcased the potential for guided synthesis, they often faced challenges related to training instability and limited output diversity. The advent of conditional VAEs~\citep{NIPS2015_8d55a249,yan2016attribute2image} initiated a notable transition toward probabilistic models. These models provided a more principled approach to managing uncertainty and enhancing sample diversity, although this sometimes impacted the quality of the generated outputs. They established critical theoretical frameworks for modeling the interactions between conditioning information and generated results. Recently, conditional diffusion models~\citep{corenflos2024conditioning,bansal2023universal, Dhariwal2021DiffusionMB, Ho2022ClassifierFreeDG, Nichol2021GLIDETP} have surfaced as the leading paradigm, fundamentally transforming expectations for conditional generation tasks. By intertwining the mathematical elegance of gradual noise reduction with advanced conditioning mechanisms, these models deliver unparalleled accuracy in converting conditioning signals—especially text descriptions—into coherent visual outputs. Our research extends the boundaries of conditional diffusion in a groundbreaking way, applying these techniques to generate neural network parameters. This marks a significant shift from conventional media-focused applications, highlighting the broader potential of conditional diffusion as a comprehensive framework for structured output synthesis across various problem domains.

\paragraph{Parameter Generation.}
The domain of parameter generation has significantly evolved, addressing the challenge of learning distributions over neural network weights. Early methods in parameter learning included stochastic neural networks~\citep{Sompolinsky1988ChaosIR, Bottou1991StochasticGL, Graves2011PracticalVI} and Bayesian neural networks~\citep{Neal1995BayesianLF, kingma2013auto, Gal2015DropoutAA}, aimed at modeling probability distributions over weights to improve generalization and uncertainty estimation. While theoretically appealing, these faced scalability issues with modern architectures. HyperNetworks~\citep{Ha2016HyperNetworks} advanced the field by generating parameters for other networks. This idea expanded in SMASH~\citep{Brock2017SMASHOM}, enhancing architecture support through memory read-write mechanisms. Recent work has utilized diffusion models for parameter generation~\citep{Li2024AutoregressiveIG, Li2024TexttoModelTN, Lin2024UnleashGN, Soro2024DiffusionbasedNN, wang2024neuralnetworkdiffusion, Erko2023HyperDiffusionGI, Peebles2022LearningTL}. Notable progress has been made in representation learning; NeRN~\citep{Ashkenazi2022NeRNL} showed that neural representations can directly encode weights of pre-trained convolutional networks, while Hyper-Representations~\citep{Schrholt2022HyperRepresentationsAG} used autoencoders to capture latent distributions. Furthermore, conditional parameter generation has emerged as a promising area. COND P-DIFF~\citep{jin2024conditionalloraparametergeneration} and Tina~\citep{Li2024TexttoModelTN} introduced text-controlled methods for parameter generation, facilitating semantic guidance of weight synthesis. Yet, many current methods struggle with large-scale architectures like ResNet, ViT, or ConvNeXt. RPG~\citep{wang2025recurrentdiffusionlargescaleparameter} proposes a tokenization strategy with recurrent diffusion for large-scale generation, but conditional generation remains limited. Our work builds on these studies, proposing a new method for large-scale conditional parameter generation for LoRA parameters that maintains high performance across various tasks.

\section{Preliminary}
\subsection{Conditional Diffusion Models}
A diffusion model~\citep{Ho2020DenoisingDP} is the state-of-the-art generation model, normally consisting of a forward process that progressively adds noise to training data, and a reverse process that learns to remove noise step by step. When conditioned on auxiliary information, these models further enable controllable or guided generation, often referred to as a conditional diffusion model. In this section we go over the basics of (conditional) diffusion models. 

\paragraph{Forward Process.} Let $x_0\in\mathcal{X}$ be a data sample drawn from the true distribution $q(x_0)$. For a total of $T$ time steps, the forward process incrementally corrupts $x_0$ into $\{x_1, x_2, \dots, x_T\}$ by adding Gaussian Noise:
\begin{equation}
    q(x_t | x_{t-1}) = \mathcal{N}(x_t;\sqrt{1-\beta_t}x_{t-1}, \beta_t I).
\end{equation}
Here $\beta_t$ is a small variance scheduled over steps $t=1,\dots,T$, and $I$ is the identity matrix. After many steps $x_T$ is nearly pure noise and thus easy to sample from a known Gaussian. Crucially, this forward process does not require training: it is defined by a simple Markov Chain. An important convenience is that at any step $t$, one can directly draw $x_t$ from $x_0$ via 
\begin{equation*}
    x_t = \sqrt{\alpha_t}x_0 + \sqrt{1 + \alpha_t}\epsilon, \; \alpha_t=\prod_{s=1}^t(1-\beta_s), \;\epsilon\sim \mathcal{N}(0,I)
\end{equation*}
This closed-form reparameterization is often used to simplify training objectives.

\paragraph{Reverse Process.} While the forward process is explicitly defined, the reverse process is more challenging: given a noisy sample $x_t$, the model must predict $x_{t-1}$. One assumes a Gaussian form:
\begin{equation}
    p_\theta(x_{t-1} | x_t) = \mathcal{N}\left(x_{t-1}; \mu_\theta(x_t, t), \Sigma_\theta(x_t,t)\right)
\end{equation}
where $\mu_\theta$ and $\Sigma_\theta$ are learned neural networks with parameters $\theta$. Training proceeds by matching this learned reverse process to the true (unknown) reverse distribution. A common simplified objective is the mean-squared error between the model's predicted noise and the actual noise injected in the forward process:
\begin{equation}
    \mathcal{L}(\theta) = \sum_{t=1}^T \mathbb{E}_{x_0, \epsilon}\left[\| \epsilon - \epsilon_\theta(x_t, t)\|^2\right]
\end{equation}
where $x_t$ is obtained by sampling noise $\epsilon$ and applying the forward formula above. Minimizing $\mathbb{L}(\theta)$ forces the reverse process to remove noise properly at each time step. At inference time, one starts from a random Gaussian sample $x_T$ and sequentially applies $p_\theta(x_{t-1}, x_t)$ to denoise it from $t=T$ down to $t=1$. The result is a generated sample $x_0$ that (approximately) follows the data distribution.

\paragraph{Conditional Diffusion Models.} To allow controllable generation, a conditional diffusion model augments each step with a condition $c$. This condition can represent any auxiliary information relevant to the task—for instance, a text prompt, a domain label, or a style embedding. The forward process typically remains unchanged:
\begin{equation}
    q(x_t | x_{t-1}, c) = q(x_ | x_{t-1})
\end{equation}
but the reverse process is now defined by 
\begin{equation}
    p_\theta(x_{t-1} | x_t, c) = \mathcal{N}\left(x_{t-1}; \mu_\theta (x_t, t, c), \Sigma_\theta(x_t, t, c)\right)
\end{equation}
We then modify the training objective so that at each step $t$, the model sees not only the noisy sample $x_t$ but also the condition $c$. A practical variant of the training loss becomes:
\begin{equation}
    \mathcal{L}_{cond}(\theta) = \sum_{t=1}^T \mathbb{E}_{x_0, \epsilon, c} \left[\|\epsilon - \epsilon_\theta(x_t, t, \tau(c) \|^2 \right]
\end{equation}
where $\tau(c)$ is an encoder or projection function that maps $c$ into a representation usable by the network. At sampling time, we repeatedly apply the reverse distribution
\begin{equation}
    x_t \sim p_\theta(x_{t-1} | x_t, c), \;\;\;\; t=T, \dots, 1
\end{equation}
ensuring that the resulting $x_0$ is guided toward the desired condition. This approach thus enables flexible, high-quality generation with explicit control over the style, content, or any other information captured by $c$.

\subsection{Low-Rank Adaptation}
Let $W_0\in\mathbb{R}^{d\times d}$ be a pretrained weight matrix in a transformer model—for example, the weight of a particular layer. A typical fine-tuning procedure would update $W_0$ directly, requiring substantial storage and computation for large-scale models. Low-Rank Adaptation or simply LoRA~\citep{Hu2021LoRALA} offers a more parameter-efficient approach by assuming that the update $\Delta W \in \mathbb{R}^{d\times d}$ to $W_0$ (the difference between $W_0$ and the fully fine-tuned matrix) can be factorized into a low-rank form:
\begin{equation}
    \Delta W = BA
\end{equation}
where $B\in \mathbb{R}^{d\times r}$ and $A\in\mathbb{R}^{r\times d}$ are trainable matrices, and $r\ll d$ is the rank of the decomposition. This drastically cuts the number of learnable parameters: instead of adjusting all $d^2$ entries of $\Delta W$, LoRA only requires tuning $2rd$ parameters (the entries in $A$ and $B$). Thus, after fine-tuning, the adapted weight matrix is
\begin{equation}
    W_{new} = W_0 + \Delta W = W_0 + BA
\end{equation}
enabling the model to learn task-specific adaptations using significantly fewer parameters. This approach maintains much of the flexibility of conventional fine-tuning while being more efficient in both computation and memory.

\section{Method}
\subsection{Overview}
In this section, we introduce \texttt{ORAL} (L\textbf{oRA} via conditiona\textbf{l}), our large-scale conditional LoRA generation framework. Building on the idea of recurrent diffusion from RPG~\citep{wang2025recurrentdiffusionlargescaleparameter}, our approach fuses two distinct condition sources: (1) base-model encoding that identify which pretrained foundation model we are adapting, and (2) textual instructions that specify the target task or style. These two conditions feed directly into a diffusion-based architecture, augmented by a recurrent module. By decomposing the LoRA matrices into tokens, we enable scalable parameter generation for hundreds of millions of weights, while retaining explicit conditional control.

\subsection{Two-Part Conditional Modules} 
We aim to synthesize LoRA updates $\Theta$ suitable for adapting a foundation model $W_0$ to a specific downstream task. Formally, a LoRA update seeks to fine-tune via a low-rank update $\Delta W = BA$, with $B\in \mathbb{R}^{d\times r}$ and $A\in \mathbb{R}^{r\times d}$. However, in practice, a large transformer typically has multiple trainable matrices across different layers; we use $\Theta = \{\Delta W^{(l)}\}$ to denote all LoRA updates, where each $\Delta W^{(l)}$ factors into $B^{(l)}A^{(l)}$. Merging these updates with $W_0$ yields the adapted weights for inference. Our approach aims to generate these updates $\Delta W$ directly via a diffusion model conditioned on: (1) the base model and (2) a textual description of the task. 

\paragraph{Base-Model Encoding.} We rely on an encoder $f_{base}$ that maps the entire base model weights $W_0$ to a compact embedding $c_{model} = f_{base}(W_0)$. We flatten certain structural metadata from $W_0$ (dimension, layer count, special tokens), and convert this into a string. Then we pass this string through an encoder which will generate an embedding vector. This embedding vector ensures that the generative process "knows" the architecture and dimension of $W_0$, so that the LoRA updates we synthesize are structurally compatible with it.

\paragraph{Textual Prompt Encoding.} We simultaneously feed in a textual description $\mathcal{T}$ of the downstream task--e.g. "LoRA adapter for sentiment classification", or "LoRA Adapter for SST-2 task" alongside some fewshot examples from the dataset itself. An encoder $f_{text}$ converts $\mathcal{T}$ into a condition vector $c_{text} = f_{text}(\mathcal{T})$. This can be realized by a pretrained text encoder like CLIP~\citep{Radford2021LearningTV} or T5~\citep{Raffel2019ExploringTL}, depending on the application. This textual embedding imposes semantic constraints on the desired LoRA adapter.

Finally, we combine these two embeddings through concatenation into a global condition $c$: 
\begin{equation}
    c = \left[c_{model}\;;\; c_{text}\right]
\end{equation}
Throughout training and inference, $c$ is provided to the generative modules so that both model encodings and textual instructions guide the LoRA generation.

\subsection{LoRA Tokenization and Recurrent Prototypes}
\paragraph{Tokenization.} To handle LoRA updates at large scale, we adopt a tokenization approach similar to RPG~\citep{wang2025recurrentdiffusionlargescaleparameter}. In particular, we break each $\Delta W^{(l)}\in \mathbb{R}^{d_l \times d_l}$ into fixed-size tokens:
\begin{itemize}
    \item[\ding{182}]\textbf{Flatten and Group}: We flatten each layer's LoRA update $\Delta W^{(l)}$ and then apply normalization per layer to reduce any distributional shifts.
    \item[\ding{183}]\textbf{Splitting into Tokens}: We slice flattened vector at a predetermined token size $k$. Large layers simply produce more tokens; small layers produce fewer, and to make them of the same lengths, we pad them accordingly. 
    \item[\ding{184}]\textbf{Positional Annotations}: Each token $u_j$ is annotated with a layer index $l$ and an intra-layer offset to preserve ordering. We represent these via simple sinusoidal embeddings $p_j$ concatenated with the tokens.
\end{itemize}
After repeating the above process for all layers, we obtain a sequence $\{u_1, \dots, u_T\}$. Each LoRA checkpoint $\Theta_i$ thus yields a "token list" plus conditions $c_i$.

\paragraph{Recurrent Prototypes.} Following the insights from RPG~\citep{wang2025recurrentdiffusionlargescaleparameter}, we incorporate a recurrent module $f_\phi$ to handle the token sequence. Given the tokens $u_j$, the module produces "prototype" vectors $p_j$, that summarize local correlation. Formally, 
\begin{equation}
    p_j, h_j = f_\phi (u_j, h_{j-1})
\end{equation}
where $h_{j-1}$ is the recurrent hidden state from token $j-1$. We use a small memory-efficient module based on Mamba~\citep{Gu2023MambaLS}. Once we reach the final token, the set of prototypes $\{p_1, \dots, p_T\}$ collectively encodes the entire LoRA adapter $\Theta$ in a compact representation. This ensures that the network can scale to large LoRA sets, without needing to flatten them into one huge input. 

\subsection{Conditional Diffusion}
We adopt diffusion-based generative model, specifically 1D convolution network, to denoise each token, conditioning on both the recurrent prototypes $p_j$ and global conditions $c$. Let $u_{j,0}=u_j$ be the original clean token. The forward process adds noise across time steps $T$:
\begin{equation}
    u_{j,t} = \alpha_t u_{j,0} + \sigma_t \epsilon, \;\;\;\; \epsilon\sim\mathcal{N}(0, I)
\end{equation}
Yielding highly corrupted tokens at large $t$. The reverse process is learned by a denoising network $\epsilon_\theta$, which predicts the noise given the current state, the prototype $p_j$, and the condition $c$. We train this network using the usual noise prediction loss: 
\begin{equation}
    \mathcal{L}_{diff}(\theta, \phi) = \sum_{t=1}^T \mathbb{E}\left[ \| \epsilon - \epsilon_\theta (u_{j,t}, p_j, c, t \|^2\right]
\end{equation}
This allows us to train the diffusion model, as well as pass the gradient through the recurrent module, without having to explicitly train it. 

\begin{table*}[t]
\centering
\hspace{-0.8em}
\subfloat[\small FID scores (lower is better) for Stable‐Diffusion 2.1 adapted to four target styles. “Stable-Diffusion 2.1” uses the original model directly, “Original LoRA” denotes standard fine‐tuning, and “Ours”. Best results are represented by \textbf{bold}.\label{tab:image_tasks}]{
\centering
\begin{minipage}[h]{0.45\textwidth}
\begin{center}
\centering
\tablestyle{5pt}{1.15}

\begin{tabular}{l| c c  c c }
    & Pokemon & PixelArt & Cartoon &Retro\\
    & FID($\downarrow$) & FID($\downarrow$) & FID($\downarrow$)& FID($\downarrow$) \\ 
\shline
    Stable-Diffusion 2.1%
    & 89.45 & 104.56 & 110.22 & 145.23 \\
    Original LoRA%
    & 24.40 & 26.89 & \textbf{25.61} & 42.39\\
    \baseline{\texttt{ORAL} (Ours)}%
    & \baseline{\textbf{23.95}} & \baseline{\textbf{26.32}} & \baseline{25.77} & \baseline{\textbf{41.50}} \\

    \end{tabular}

\end{center}
\end{minipage}
}
\hspace{1em}
\subfloat[\small Performance comparison on Flickr30K, NoCaps, and DocVQA datasets. Retrieval@1 is reported for Flickr30K and NoCaps, and accuracy is reported for DocVQA. Best results are in \textbf{bold}.\label{tab:multimodal}]{
\centering
\begin{minipage}[h]{0.45\textwidth}
\begin{center}
\tablestyle{4pt}{1.15}
\begin{tabular}{l| c c  c}
    & Flickr30K & NoCaps & DocVQA \\
    & Retrieval@1($\uparrow$) & Retrieval@1($\uparrow$) & Accuracy($\uparrow$) \\ 
\shline
    Qwen-7B-VL%
    & 85.78 & 121.42 & 65.11 \\
    Original LoRA%
    & 96.65  & 132.76 & \textbf{84.53}\\
    \baseline{\texttt{ORAL} (Ours)}%
    & \baseline{\textbf{96.72}} & \baseline{\textbf{134.81}} & \baseline{84.49}\\

    \end{tabular}

\end{center}
\end{minipage}}

\vspace{-0.5em}
\caption{\small Experimental results on different Image Generation and Multi-modal Tasks.}
\label{tab:exp1}
\end{table*}

\section{Experiments}

\begin{figure}
    \centering\scalebox{0.80}{
    \includegraphics[width=0.4\textwidth]{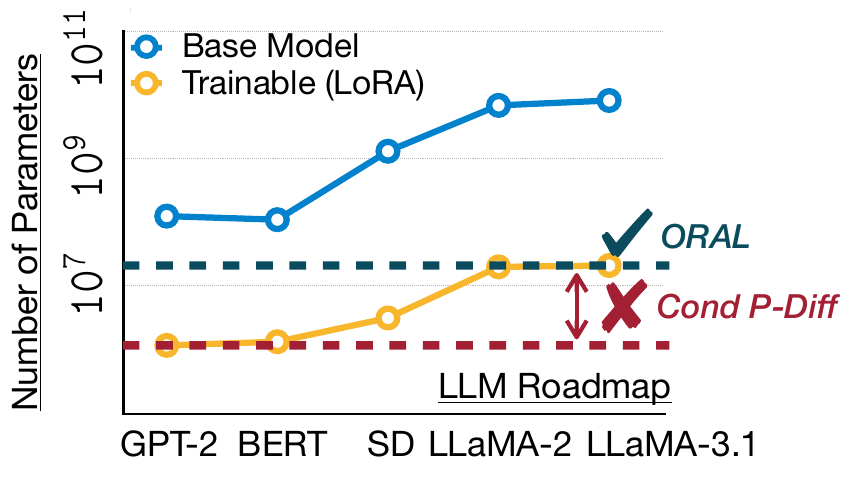}
}
    \vspace{-4mm}
    \caption{\small Comparison of parameter generation capacity between our proposed method, \texttt{ORAL}, and the baseline Cond P-Diff. \texttt{ORAL} effectively generates LoRA adapters at a significantly larger scale (e.g., 7B Models), surpassing the capacity of Cond P-Diff, which fails to operate efficiently at higher parameter scales. This demonstrates \texttt{ORAL}'s ability to handle large-scale parameter synthesis, crucial for adapting modern large language models.}
    \label{fig:comp}
    \vspace{-3mm}
\end{figure}

\begin{table*}[h]
\centering
\tablestyle{6pt}{1.2}
\begin{tabular}{l| l l l l l l l }
    & BoolQ & SST-2 & MRPC & RTE & Winogrande & WNLI & GSM8K \\
    & Accuracy($\uparrow$) & Accuracy($\uparrow$) & Accuracy($\uparrow$) & Accuracy($\uparrow$) & Accuracy($\uparrow$) & Accuracy($\uparrow$) & Accuracy($\uparrow$)\\ 
\shline
    Mistral-7B%
    & 83.58 & 66.86 & 65.20 & 67.51 & 74.11 & 57.76 & 6.37 \\
    Original LoRA%
    & 91.01  & 95.99 & 89.46  & 87.73 & 85.95 & 83.11 & 32.04\\
    \baseline{\texttt{ORAL} (Ours)}%
    & \baseline{90.67 \textcolor{red}{$\downarrow 0.25$ }} & \baseline{96.01 \textcolor{green}{$\uparrow 0.02$ }} & \baseline{89.23 \textcolor{red}{$\downarrow 0.23$ }} & \baseline{86.34 \textcolor{red}{$\downarrow 1.39$ }} &
    \baseline{86.10 \textcolor{green}{$\uparrow 0.15$ }} & \baseline{83.11 \textcolor{green}{$\uparrow 0.00$ }} & 
    \baseline{34.67 \textcolor{green}{$\uparrow 0.63$ }}\\

    \end{tabular}
\caption{\small Performance comparison on seven NLP datasets (BoolQ, SST-2, MRPC, RTE, Winogrande, WNLI, GSM8K) using accuracy as the evaluation metric. We report results for the Mistral-7B base model, Original LoRA baseline, and our conditional recurrent diffusion method. }
\vspace{-2mm}
\label{tab:exp3}
\end{table*}

In this section, we will evaluate our conditional large-scale LoRA generation method to generate parameters across multiple tasks, models, and modalities. We first report our experimental setup that is constant across all experiments, like training details, and then experiment-dependent settings and results are contained within their own subsections. 

\subsection{Experimental Setup}

\paragraph{Training Details.} We train our conditional recurrent diffusion model for $100,000$ iterations on eight A6000 GPUs. Each run processes tokens of fixed size 8196, leveraging a Mamba~\citep{Gu2023MambaLS} block as the recurrent backbone alongside a 1D convolution diffusion network. We encode the base model via a text description passed through a BERT-Base-Uncased~\citep{DBLPabs-1810-04805} encoder, while task prompts are handled by the CLIP~\citep{Radford2021LearningTV}. Within each experiment, we devote one training pipeline per task, ensuring the model fully learns the distribution of LoRA parameters tailored to that specific objective. For all the experiments we set our LoRA rank to be $8$, however, an ablation study on various ranks was done.

\paragraph{Inference Details and Comparisons.} During inference, our approach takes as input both the textual prompt and the base‐model description. We initialize each token with Gaussian noise and then run through our learned 1D convolutional diffusion network, conditioned on the recurrent prototypes from Mamba and the global conditions. Thanks to the tokenization scheme, inference scales to hundreds of millions of LoRA parameters without memory overflow. Notably, no existing method can generate large‐scale LoRA parameters conditionally in this manner, so we do not compare against external baselines for a like‐for‐like evaluation. Instead, we focus on our method’s ability to effectively synthesize LoRA adapters across various tasks and model sizes and compare them to the original results. For evaluation, we synthesize LoRA parameters three times (via different random seeds) for each target condition. We merge these generated LoRAs into the base model and measure downstream performance, reporting the average (and sometimes best/min) accuracy.

\begin{figure*}[!htbp]
    \centering\scalebox{0.80}{
    \includegraphics[width=1\textwidth]{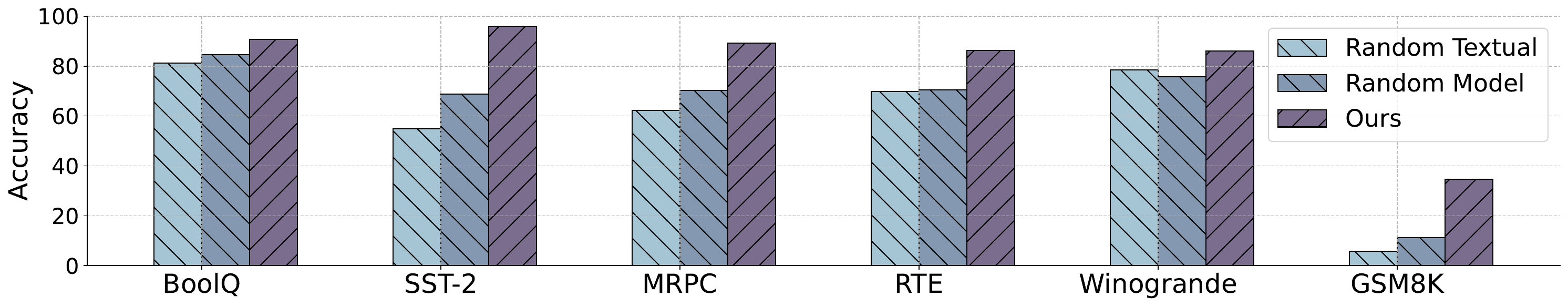}
}
    \vspace{-4mm}
    \caption{\small Ablation results showing accuracy comparisons across NLP tasks using random model embeddings, random textual embeddings, and our method with meaningful embeddings. Higher accuracy achieved by our conditional embeddings highlights their importance in guiding effective LoRA adapter generation.}
    \label{fig:ablation1}
    \vspace{-3mm}
\end{figure*}   

\subsection{Exp 1: Performance on Image Generation Task}
\paragraph{Setup and Datasets.} In this experiment, we test our method's ability to adapt Stable-Diffusion 2.1~\citep{rombach2022high}, to four distinct styles, each realized as a LoRA Adapter: Pokemon, PixelArt, Cartoon, and Retro. Concretely, we treat Stable-Diffusion as the base model and synthesize separate LoRA parameters to specialize it for each of the four styles. We evaluate image generation quality using the FID score, sampling images from the adapted model in each style and comparing their statistics to a style-specific reference set. We compare the results from our generated LoRA with the original LoRA as well as the zero-shot performance of the base model to motivate the need for fine-tuning.

\paragraph{Results.} Table~\ref{tab:image_tasks} reports the FID scores for all four tasks and we can summarize our results as follows: \ding{182}\underline{\textbf{Comparison with original LoRA:}} Our method closely matches or slightly surpasses standard LoRA fine-tuning in terms of style fidelity, as measured by the FID. In particular, we observe improvements in FID for Pokemon ($0.45$ decrease) and PixelArt ($0.57$ decrease), while achieving nearly identical results for Cartoon ($0.16$ increase) and a clear improvement for Retro ($0.89$ decrease). These results demonstrate that our conditional recurrent diffusion method successfully captures and reproduces most of the style-specific characteristics traditionally obtained through gradient-based fine-tuning. \ding{183}\underline{\textbf{Comparison with Base Model:}}Compared to the original Stable-Diffusion 2.1 (without LoRA), our generated LoRA adapters significantly enhance image-generation quality across all evaluated styles. Specifically, we achieve substantial FID reductions for the evaluated styles—Pokemon ($89.45$ $\to$ $23.95$), PixelArt ($104.56$ $\to$ $26.32$), Cartoon ($110.22$ $\to$ $25.77$), and Retro ($145.23$ $\to$ $41.50$). These improvements confirm our model’s effectiveness at synthesizing style-specific adapters, greatly enhancing style consistency.

\vspace{-2mm}
\subsection{Exp 2: Performance on Multi-modal Tasks}
\vspace{-1mm}
\paragraph{Setup and Datasets.} The goal of this experiment is to evaluate our framework for LoRA parameter generation on multi-modal tasks. The experimental setup consists of adapting the base model Qwen-7B-VL~\citep{Qwen-VL}, using LoRA adapters specialized for three distinct multimodal tasks: (1) Flickr30K~\citep{Plummer2015Flickr30kEC}: a captioning task, (2) NoCaps~\citep{Agrawal2019nocapsNO}: another captioning task, and (3) DocVQA~\citep{Mathew2020DocVQAAD}: a document based Visual Questioning Answering Task. We compare our method against the zero-shot performance of the base model, as well as the original LoRA adapter to showcase the generation. For Flickr30K and NoCaps, we utilize the Image-to-Text Retrieval at 1 metric, while for the DocVQA task we use Accuracy.

\paragraph{Results.} The results for all three multi-modal tasks are summarized in Table~\ref{tab:multimodal}. Our method closely matches or slightly surpasses the Original LoRA baseline for image-to-text retrieval tasks, as measured by Retrieval@1. Specifically, we observe improvements in Retrieval@1 for Flickr30K ($0.07$ increase) and NoCaps ($2.05$ increase). However, for the DocVQA dataset, our accuracy slightly trails Original LoRA by $0.04\%$. These results confirm that our conditional recurrent diffusion method effectively captures task-specific characteristics, demonstrating comparable or superior performance to gradient-based LoRA fine-tuning.

\begin{table*}[h]
\centering
\tablestyle{6pt}{1.2}
\begin{tabular}{c|cc|cc|cc|cc|cc}
\multirow{2}{*}{Dataset}&  \multicolumn{2}{c|}{$R=2$} & \multicolumn{2}{c|}{$R=4$} & \multicolumn{2}{c|}{$R=8$} & \multicolumn{2}{c|}{$R=16$} & \multicolumn{2}{c}{$R=32$}\\
& Original & Ours 
& Original & Ours
& Original & Ours 
& Original & Ours 
& Original & Ours \\
\shline
BoolQ &90.52 & 91.43 & 90.81 & 91.27 & 91.01 & 90.76 & 91.07 & 90.74 & 90.00 & 88.44\\
MRPC & 87.30 & 88.42 & 89.42 & 89.96 & 89.46 & 89.23 & 89.22 & 88.99 & 88.97 & 87.04\\
\end{tabular}
\caption{\small  Performance comparison between Original LoRA and our method on BoolQ and MRPC datasets across different LoRA ranks R. Results are reported in accuracy (\%).}
\label{tab: ablation2}
\end{table*}

\subsection{Exp 3: Performance on NLP Tasks}
\paragraph{Setup and Datasets.} We would also like to evaluate the quality of our LoRA adapters for NLP Tasks. We test our framework's ability to adapt the base model, Mistral-7B~\citep{Jiang2023Mistral7}, using LoRA adapters specifically tailored for seven diverse NLP datasets where five of them: BoolQ (Boolean Question-Answering Task), SST-2 (Sentiment Analysis Task), MRPC (Paraphrase Detection Task), RTE (Recognizing Textual Entailment Task), WNLI (Natural Language Inference Task) are from SuperGLUE Benchmark~\citep{Wang2019SuperGLUEAS} as well as 2 other tasks: Winogrande Benchmark~\citep{Sakaguchi2019WinoGrande} (Commonsense Reasoning Task), and GSM8K~\citep{cobbe2021gsm8k} (Grade-school Math Problems). For each downstream task, we calculate the accuracy and report it for comparison. 

\paragraph{Results.} Our third experiment evaluates our method's effectiveness in generating LoRA parameters for natural language processing tasks using the Mistral-7B model. As shown in Table~\ref{tab:exp3}, our generated LoRA parameters demonstrate strong performance across seven diverse NLP tasks. When compared to the Original LoRA baseline, our method achieves competitive or superior results, with slight improvements on SST-2 ($+0.02\%$), Winogrande ($+0.15$), and GSM8K ($+0.63$), identical performance on WNLI, and only marginally lower accuracy on BoolQ ($-0.25$), MRPC ($-0.23$), and RTE ($-1.39$). More importantly, our generated parameters substantially outperform the Mistral-7B base model across all tasks, with remarkable improvements ranging from $+11.99$ on Winogrande to $+29.15$ on SST-2 and $+28.30$ on GSM8K. The most significant gains appear in specialized reasoning and classification tasks, demonstrating that our conditional generation approach can effectively capture task-specific adaptations without traditional gradient-based fine-tuning. These results validate our proposed method's capability to generate high-quality LoRA parameters that match or occasionally exceed traditional fine-tuning approaches while offering the advantages of our scalable, conditional generation framework.

\vspace{-2mm}
\subsection{Exp 4: Generalization on Evolving Models}
\vspace{-2mm}

\begin{figure}
    \centering
    \vspace{-5mm}
    \begin{minipage}[b]{0.42\textwidth}
        \centering
        \includegraphics[width=\textwidth]{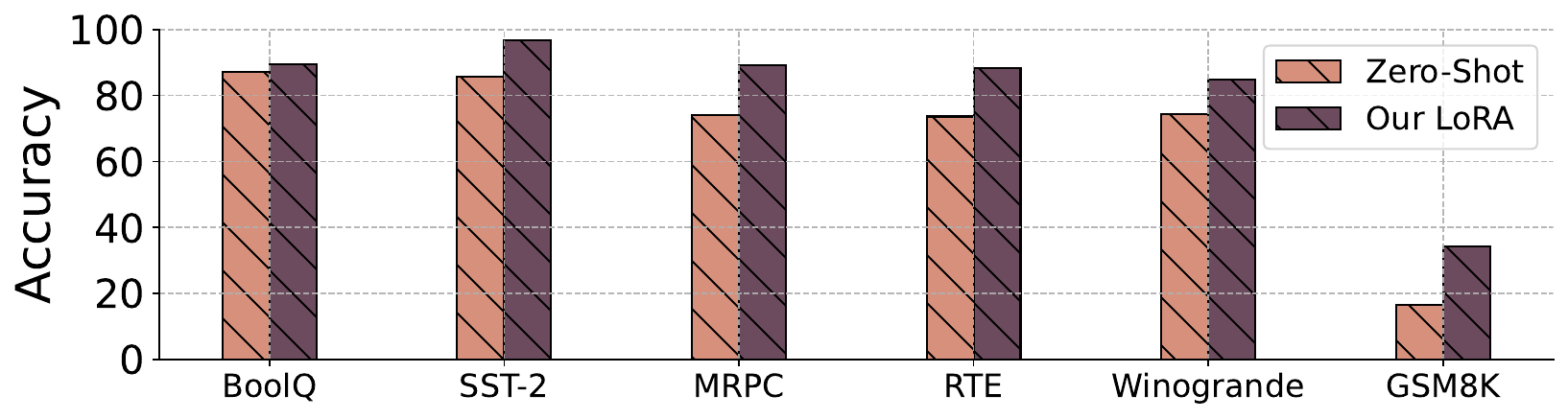}
        \caption{\small Accuracy comparison of our synthesized LoRA adapters against zero-shot base models on the unseen evolved Mistral continually pretrained on AlpacaGPT4.}
        \label{fig:alpaca}
    \end{minipage}
    \hfill
    \begin{minipage}[b]{0.42\textwidth}
        \centering
        \includegraphics[width=\textwidth]{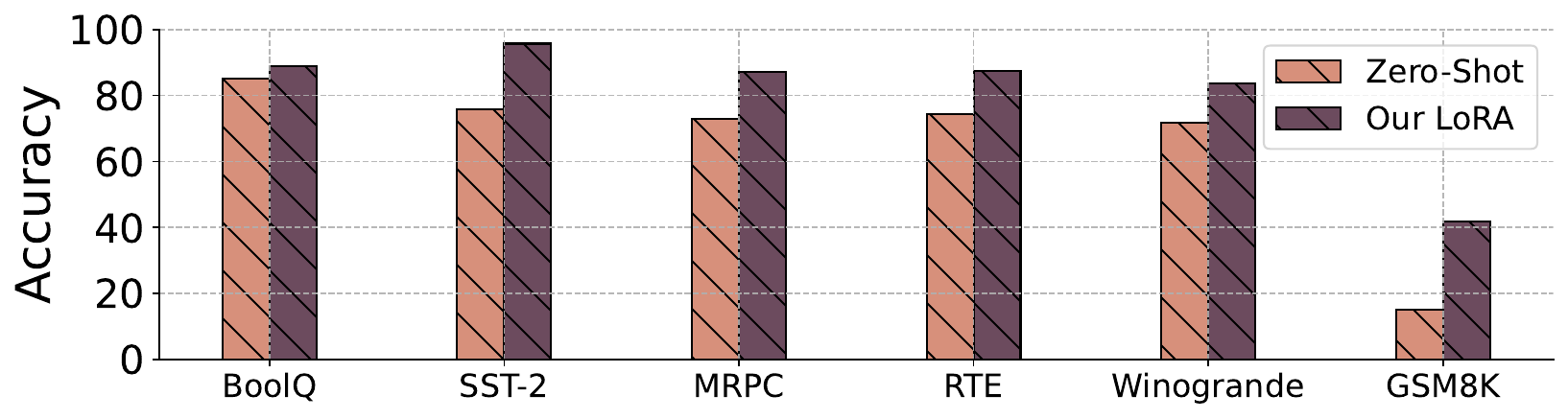}
        \caption{\small Accuracy comparison of our synthesized LoRA adapters against zero-shot base models on the unseen evolved Mistral continually pretrained on GPT4LLM.}
        \label{fig:gpt}
    \end{minipage}
    \vspace{-15pt}
\end{figure}
\paragraph{Setup and Datasets.} In this experiment, we investigate the generalization capability of our conditional recurrent diffusion framework to synthesize LoRA Adapters for evolving base models not seen during training. PortLLM~\citep{khan2024portllm} shows that most of the closed source models go through data updates (i.e., evolution) that changes the model weights, hence requiring re-fine-tuning, which can be quite costly sometimes, or can be a hindrance due to the lack of data availability (time-sensitive access). To tackle this, we showcase the generalizability of our method in such an evolving setting, so that we do not have to re-fine-tune a newer evolved model on a downstream task, instead we can prompt our framework to generate the LoRA parameters that can be directly plugged in any iteration of the base model. Specifically, we use LoRA adapters trained on base models of the same architecture but different pertaining data to simulate the evolution. For the downstream tasks we utilize six datasets: BoolQ, SST-2, MRPC, RTE, Winogrande and GSM8K, and for the pretrained base model we utilize the 3 different time steps (evolutions): Base Mistral-7B Model ($t=0$), Mistral continually pretrained on \{OpenOrca\}~\citep{OpenOrca} ($t=1$), and Mistral continually pretrained on \{OpenOrca,OpenPlatypus~\citep{platypus2023}\} ($t=2$). For each of these base models, we curate all 6 LoRAs, so we have a total of 18 LoRAs in our training set. Then we synthesize LoRA adapters for $t=3$ and $t=4$ for base models: Mistral continually pretrained on \{OpenOrca, OpenPlatypus, AlpacaGPT~\citep{alpaca}\} and Mistral continually pretrained on \{OpenOrca, OpenPlatypus, AlpacaGPT, GPT4LLM~\citep{peng2023instruction}\}, which are both unseen base models for the trained pipeline. Then we compare the zero-shot performance of these pretrained models, against our generated LoRAs to showcase the improvement.

\paragraph{Results.} The results for the experiment are summarized in Figures~\ref{fig:alpaca} and~\ref{fig:gpt}, comparing zero-shot performance on two unseen evolved base models (AlpacaGPT4 and GPT4LLM) against LoRA adapters generated by our framework \texttt{ORAL}. \ding{182}\underline{\textbf{Performance Improvement over zero-shot base models:}} Our generated LoRA adapters demonstrate notable improvements, with the highest accuracy increases observed on SST-2 ($10\%$) and GSM8K ($30\%$) tasks. \ding{183}\underline{\textbf{Generalization Capability:}} The synthesized LoRA adapters effectively generalize to unseen evolved models ($t=3$ and $t=4$), significantly outperforming the zero-shot performance of base models. The results demonstrate that our conditional recurrent diffusion framework successfully transfers learned adaptation knowledge to new evolving base models without requiring additional fine-tuning, underscoring the robust generalization capabilities of our method. \ding{184}\underline{\textbf{Consistency Across Models and Tasks:}} The synthesized adapters consistently demonstrate improved or comparable performance across diverse NLP tasks. Particularly, the task-specific synthesized adapters for BoolQ, SST-2, MRPC, and GSM8K consistently outperform the baseline, indicating the broad applicability of \texttt{ORAL}.

\vspace{-1mm}
\subsection{Ablation Study 1: Effect of Random Embeds.}
In this ablation study, we evaluate the impact of random embeddings on our conditional recurrent diffusion framework by comparing performance across six NLP datasets: BoolQ, SST-2, MRPC, RTE, Winogrande, and GSM8K. Specifically, we compare the accuracy achieved by our method using meaningful conditional embeddings against two baseline scenarios—random model embeddings and random textual embeddings. The results are summarized in Figure~\ref{fig:ablation1}.

From the results, we observe that using meaningful embeddings consistently outperforms both random embedding scenarios across all evaluated datasets. This indicates the importance of meaningful conditioning for synthesizing high-quality LoRA adapters. Interestingly, we see a more substantial performance drop with random textual embeddings than with random model embeddings, particularly noticeable in tasks requiring nuanced semantic understanding, such as GSM8K and SST-2. For example, in GSM8K, random textual embeddings show substantially lower accuracy than our method, underscoring the critical role of textual conditioning in generating high-quality, task-specific LoRA adapters. Conversely, random model embeddings still demonstrate moderate performance, suggesting that while base-model conditioning contributes to effective parameter generation, textual conditioning provides more crucial guidance.

\subsection{Ablation Study 2: Effect of LoRA Rank}
In this ablation study, we analyze the impact of varying the LoRA rank (R) on the quality of the adapters generated by our conditional recurrent diffusion method. We compare the performance across two NLP datasets: BoolQ and MRPC, using LoRA ranks of $2$, $4$, $8$, $16$, and $32$.

The results in Table~\ref{tab: ablation2} provide some interesting insights: \ding{182}\underline{\textbf{Comparison with Original LoRA:}} Our method consistently matches or surpasses the performance of Original LoRA across all ranks, showing the effectiveness of conditional generation even with limited parameter capacity. \ding{183}\underline{\textbf{Performance Variation Across LoRA Ranks:}} We observe that as the LoRA rank increases, performance initially improves and then slightly declines. For example, on BoolQ, our method achieves the highest accuracy at rank $R=4 $($91.27\%$) but then sees a slight drop at higher ranks ($90.76\%$ for $R=8$ and further decreases at $R=32$). Similar trends are observed on MRPC, with peak accuracy at rank $4$. These observations highlight the importance of carefully choosing an optimal LoRA rank, balancing parameter efficiency and performance. Lower ranks appear sufficient to achieve competitive performance.

\vspace{-3mm}
\section{Conclusion}
\vspace{-2mm}
We introduced \texttt{ORAL}, a novel framework for conditional recurrent diffusion that enables scalable generation of Low-Rank Adaptation~(LoRA) parameters at large scale. Our approach is designed to address the critical challenge of adapting to constantly evolving large language models without the computational burden of retraining. By incorporating both model architecture and textual task specifications as conditional inputs to the diffusion process, \texttt{ORAL} achieves remarkable flexibility and portability across model architectures while maintaining performance comparable or superior to traditionally trained LoRA parameters.

Our comprehensive experiments across various diverse tasks, three state-of-the-art pre-trained models~(Stable-Diffusion2.1, Mistral, Qwen-7B-VL), and two modalities demonstrate the practical viability of our approach for real-world deployment. \texttt{ORAL} successfully scales to generate hundreds of millions of parameters while preserving adaptation quality. In the future, exploring the interpretability of generated parameters could provide insights into how different tasks influence parameter properties.  
{
    \small
    \bibliographystyle{ieeenat_fullname}
    \bibliography{main}
}

\end{document}